\documentclass[conference]{IEEEtran}
\usepackage{color}
\usepackage{multirow}
\usepackage{array}
\usepackage{booktabs}
\usepackage{amsfonts}
\usepackage{amsmath}
\usepackage{cite}
\usepackage[pdftex]{graphicx}
\usepackage[none]{hyphenat}   
\usepackage{epstopdf}

\IEEEoverridecommandlockouts

\begin{document}
\title{\LARGE \bf LIDAR-based Driving Path Generation \\ Using Fully Convolutional Neural Networks}

\author{Luca Caltagirone$^{*}$, Mauro Bellone, Lennart Svensson, Mattias Wahde\\
	\thanks{$^{*}$Corresponding author: luca.caltagirone@chalmers.se. L. Caltagirone, M. Bellone, and M. Wahde are with the Adaptive Systems Research Group, Applied Mechanics Department, Chalmers University of Technology, Gothenburg, Sweden. L. Svensson is with the Signal and Systems Department, also at Chalmers University of Technology.
	}}

\maketitle

\begin{abstract}
In this work, a novel learning-based approach has been developed to generate driving paths by integrating LIDAR point clouds, GPS-IMU information, and Google driving directions. The system is based on a fully convolutional neural network that jointly learns to carry out perception and path generation from real-world driving sequences and that is trained using automatically generated training examples. Several combinations of input data were tested in order to assess the performance gain provided by specific information modalities. The fully convolutional neural network trained using all the available sensors together with driving directions achieved the best MaxF score of 88.13\% when considering a region of interest of $\textbf{60}\times\textbf{60}$ meters. 
By considering a smaller region of interest, the agreement between predicted paths and ground-truth increased to 92.60\%. The positive results obtained in this work indicate that the proposed system may help fill the gap between low-level scene parsing and behavior-reflex approaches by generating outputs that are close to vehicle control and at the same time human-interpretable.
\end{abstract}

\IEEEpeerreviewmaketitle

\begin{figure*}[h!]
	\centering
	\includegraphics[width=2\columnwidth]{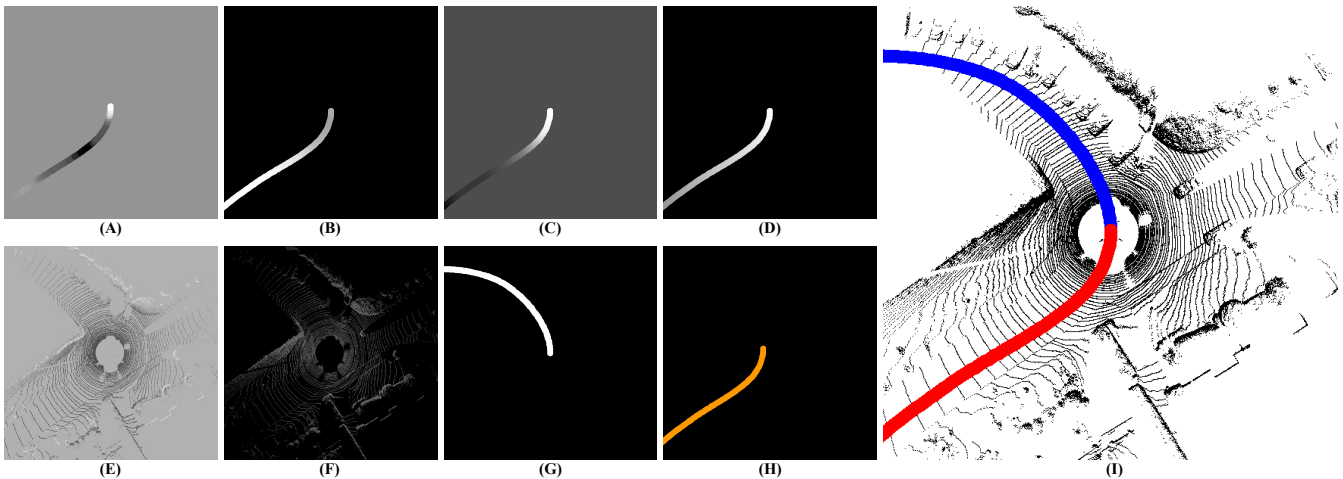}
	\caption{Overview of the input and output tensors relative to an example of the validation set (see Section~\ref{sect:dataAndTraining}). (A) Forward acceleration. (B) Forward speed. (C) Yaw rate. (D) Intention proximity. (E) Max elevation. (F) Average reflectivity. (G) Future path ground-truth. (H) Intention direction.  (I) Overlay of occupancy LIDAR image with ground-truth past path (red region) and future path (blue region). The gray-scale intensities in the sub-figures (A)--(F) are proportional to the numerical values of the corresponding illustrated quantities; the background color corresponds to zero. The possible intention directions are turn right, straight, and turn left and are represented by the color magenta, white, and orange respectively.}
	\label{fig:overviewIO}
\end{figure*}

\section{Introduction}
In recent years, universities, high-tech companies, and the automotive industry have invested vast resources into developing the technology necessary to enable fully autonomous driving. There are several reasons behind this collective effort; to name a few, it is argued that with autonomous vehicles there will be fewer accidents, less pollution, and a more efficient use of the infrastructure. 

Currently, two main paradigms exist to address the problem~\cite{ChenEtAl2015}: \textit{mediated perception} and \textit{behavior reflex}. Approaches utilizing mediated perception divide autonomous driving into subtasks, such as lane marking detection, vehicle detection, free road estimation, and traffic sign recognition. The results of each subtask are then combined to produce a world model that is afterwards analyzed to decide what driving action should be carried out.
This paradigm is currently the leading one in both industry and academia thanks to its important strengths of being modular and interpretable. Modularity makes it possible to substitute a subcomponent (e.g., a lane marking detector) for a better one whenever it becomes available, without having to redesign the entire system. Interpretability means that the output of each subcomponent can be easily understood and debugged: For example, one could look at the lane marking detections and immediately notice if the system is misunderstanding road barriers or shadows for lane markings.

The behavior reflex paradigm was first successfully demonstrated in 1989 by Pomerlau~\cite{Pomerleau1989}  who trained a simple three-layer fully connected artificial neural network, ALVINN, to predict the vehicle heading direction from camera and laser range finder data. More recently, Bojarski et al.~\cite{BojarskiEtAl2016} have used modern deep learning~\cite{LecunEtAl2015} techniques and trained a convolutional neural network (CNN) to infer appropriate steering angles given as input only forward-looking camera images. In~\cite{XuEtAl2016}, the authors proposed a more sophisticated architecture to perform driving action prediction that combines a fully convolutional neural network (FCN) for visual feature extraction with a long-short-term-memory (LSTM) recurrent neural network for temporal fusion of visual features and past sensory information.

While mediated perception approaches require time-consuming and expensive hand-labeling of training examples (e.g., selecting all the pixels belonging to lane markings in a given image), a behavior reflex system in its simplest form may only require  on-board steering angle logs and their corresponding time-stamped camera images, both of which are easily obtainable. 
However, this kind of approach works as a black-box and modularity is lost in favor of a monolithic system that maps raw input information to control actions. It is therefore difficult to understand why a system is choosing one action over another, and consequently how to correct undesired behaviors.

Chen et al.~\cite{ChenEtAl2015} have recently proposed an alternative approach, called \textit{direct perception}, that takes an intermediate position between mediated perception and behavior reflex. Their main idea is to train a CNN to map camera images to a predefined set of \textit{descriptors} such as heading angle and position in the ego-lane. It is argued that such a set provides a compact but complete representation of the vehicle's surrounding that can then be used for choosing appropriate control actions. Their approach, however, was developed within a simple driving simulator and it would probably not generalize well when applied to much more complex real-world scenarios. 

The approach described in this work also occupies an intermediate position between the two main paradigms previously described. By taking as input LIDAR point clouds, past GPS-IMU information, and driving directions, our system generates as output \textit{driving paths} in the vehicle reference frame. This is accomplished by implicitly learning, from real-world driving sequences, which regions of a point cloud are drivable and how a driver would navigate them. One of the novelties of our approach is that GPS-IMU data and driving directions are transformed into a spatial format that makes possible the direct information fusion with LIDAR point clouds.
In comparison with behavior reflex methods,  the proposed approach preserves the decoupling between perception and control while, at the same time, producing interpretable results. Whereas mediated perception methods carry out low-level scene parsing, our system generates a more abstract output that is closer to vehicle control and its training data is obtained automatically without the need of time-consuming hand-labeling.

The paper is organized as follows: In Section~\ref{sect:overview}, an overview of the proposed system is presented and it is followed by a description of the preprocessing steps applied to the raw input data. The FCN architecture is presented in Section~\ref{sect:architectures}. The data set and details about the training procedure are described in Section~\ref{sect:dataAndTraining}. The results are presented in Section~\ref{sect:experiments} and are followed by the conclusions and future work in Section~\ref{sect:conclusion}. 

\section{Driving path generation}
\label{sect:overview}
The goal of this work is to develop a system able to generate driving paths using as input LIDAR point clouds, GPS-IMU information, and driving intention. 
The problem is cast as a binary pixel-level semantic segmentation task using a fully convolutional neural network (FCN).
Since FCNs handle data structured in the form of multi-dimensional arrays (i.e., tensors), some preprocessing steps are required in order to transform the input data into a suitable format. Similar to \cite{CaltagironeEtAl2017}, here this is accomplished by considering a top-view perspective where the point
cloud is first partitioned within a discretized grid and then transformed into a 3D tensor by computing several basic statistics in each grid cell.
The GPS coordinates are used for generating the ground-truth paths followed by the vehicle and for determining the driving intention whenever an intersection is approached. Each path is also augmented with IMU information about the vehicle's forward speed, forward acceleration, and yaw rate, and then transformed into a  tensor.
Finally, the LIDAR and GPS-IMU tensors are stacked and given as input to an FCN that is trained to generate a confidence map assigning to each grid cell a probability that the vehicle will drive over that area during its future motion.

\subsection{LIDAR point cloud preprocessing}
\label{sect:lidar_images}
Vision-based algorithms for automotive applications usually work in one of two perspective spaces: camera or top-view. 
The former provides a view of the environment as seen from behind the windshield, whereas the latter offers a view of the vehicle's surroundings as seen from above, as if it were being observed from a bird flying above the scene.
A top-view perspective is, in our opinion, a natural choice to perform path generation and therefore it will be used in this work. 

The procedure to transform a raw point cloud into an input tensor for a CNN consists of several steps.
Initially, a discrete grid covering a region of interest (RoI) of $60\times60$ meters is created in the LIDAR $x$\nobreakdash-$y$ plane. 
The grid is centered in the LIDAR coordinate system and it has cells of $0.10\times0.10$ meters. 
Each point is then assigned to its corresponding grid cell, via projection on the $x$\nobreakdash-$y$ plane.
Afterwards, four basic statistics are computed in each cell: number of points, average reflectivity, minimum and maximum point elevation. 
Finally, a 2D tensor is created for storing each one of the previously mentioned statistics. 
The final output of this procedure is therefore a 3D tensor of size $4\times600\times600$.  Figure~\ref{fig:overviewIO} (E) and (F) illustrate, respectively, an example of point cloud maximum elevation and average reflectivity.

\subsection{GPS-IMU data preprocessing}
\label{sect:gpsimu_images}
As previously mentioned, the GPS information is used to generate the driving paths followed by the vehicle. 
Considering a generic \textit{current} time step $k$ of a driving sequence, the corresponding path $\Pi$, centered in the current position $\textbf{p}_{k} = [x_{k}, y_{k},z_{k}]$, is given by the union of two sets of 3D points, $\Pi^{-}$ and $\Pi^{+}$, expressed in the LIDAR coordinate system. 
Here, $\Pi^{-}$ denotes the \textit{past sub-path} which is defined in the time interval $[0:k]$; whereas  $\Pi^{+}$ is the \textit{future sub-path} and it is defined in the time interval $[k:N]$, where $N$ denotes the total number of time steps in the driving sequence.
Additionally, the past path $\Pi^{-}$ is also augmented with information provided by the IMU, specifically, forward speed $v$, forward acceleration $a$, and yaw rate $\omega$. By denoting the augmented descriptor with $\textbf{x} = [x, y, z, v, a, \omega]$, the  past path then becomes $\Pi^{-} = \{\textbf{x}_{0}, \ldots, \textbf{x}_{k}\}$.

Given that the paths are expressed as sets of points in the LIDAR coordinate system, they can also be transformed into tensors using a similar procedure to the one described in Section~\ref{sect:lidar_images}.  In this case, however, the vehicle's trajectory information is considered instead of the point elevation and the reflectivity statistics. Another difference is that it is necessary to add \textit{continuity} and \textit{thickness} to the paths which are, so far, a simple collection of discrete points.
Continuity is obtained by joining neighboring points to form a curve; then, by considering that the vehicle's width is about 1.80 meters, the curve is expanded 0.90 meters on each side in order to approximately cover the actual driving corridor. It is necessary to consider a driving corridor instead of a curve in order to provide the FCN with enough positive examples at training time. After inference, the 1D path can be recovered by finding the central curve of the FCN's output.

To summarize, the above procedure is used to generate two tensors: the first one contains information about the vehicle's past motion up to the current time step and it has a size of $3\times600\times600$. 
This tensor will be stacked with the LIDAR tensor, and will be part of the input for the FCN. 
The second tensor is the ground-truth future path that the FCN will be trained to predict as output.
Figure~\ref{fig:overviewIO} (A), (B), and (C) show an example of  forward acceleration, forward speed, and yaw rate tensors, respectively; whereas, Figure~\ref{fig:overviewIO}(G) illustrates the corresponding future path ground-truth.

\begin{figure*}[h]
	\centering
	\includegraphics[width=2\columnwidth]{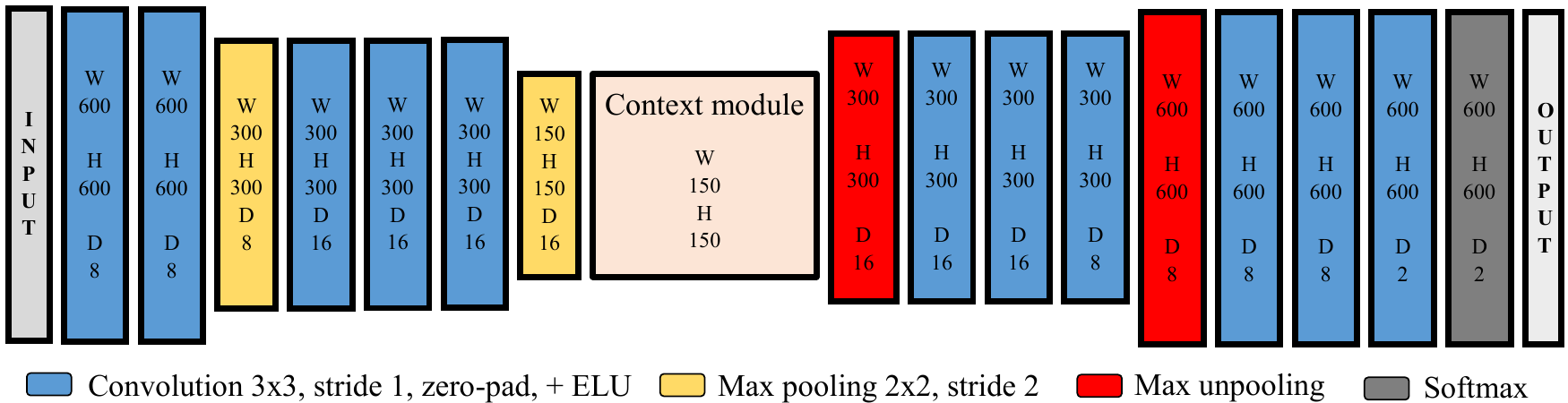}
	\caption{A schematic illustration of the proposed FCN. The input consists of a 3D tensor obtained by stacking the LIDAR and the augmented past path tensors. W represents the width, H denotes the height, and D is the number of feature maps. The FCN uses the exponential linear unit (ELU) activation function after each convolutional layer. See Table~\ref{table:nets_properties} for details about the context module architecture.}
	\label{fig:network}
\end{figure*} 

\begin{table*}[h]
	\centering
	\caption{Context module architecture. The context module consists of 13 dilated convolution layers with kernels of size $3\times3$. Each layer is followed by an ELU activation function. Layers 1-12 are also followed by a spatial dropout layer with $p_{d}=0.20$. Zero-padding is applied throughout the context module in order to preserve the width and height of the feature maps.} 
	\label{table:nets_properties}
	\begin{tabular}{|c|c|c|c|c|c|c|c|c|c|c|c|c|c|c|c|}
		\hline
		\multicolumn{1}{|c|}{Layer}&\multicolumn{1}{c|}{1}&\multicolumn{1}{c|}{2}&\multicolumn{1}{c|}{3}&\multicolumn{1}{c|}{4}&\multicolumn{1}{c|}{5}&\multicolumn{1}{c|}{6}&\multicolumn{1}{c|}{7} &\multicolumn{1}{c|}{8} &\multicolumn{1}{c|}{9}&\multicolumn{1}{c|}{10}&\multicolumn{1}{c|}{11}&\multicolumn{1}{c|}{12}&\multicolumn{1}{c|}{13}\\
		\hline
		Dilation W & 1 & 1 & 1 & 2 & 4 & 8 & 12 & 16  & 20  & 24 & 28 & 32 & 1 \\
		Receptive field W & 3 & 5 & 7 & 11 & 19 & 35 & 59 & 91 & 131 & 179 & 235 & 299 & 301\\
		\hline
		Dilation H & 1 & 1 & 2 & 4 & 8 & 12 & 16 & 20 & 24  & 28 & 32 & 1 & 1 \\
		Receptive field H & 3  & 5 & 9 & 17 & 33 & 57 & 89 & 129 & 177 & 233 & 297 & 299 & 301\\
		\hline
		\# Feature maps & 96 & 96 & 96 & 96 & 96 & 96 & 96 & 96 & 96 & 96 & 96 & 96 & 16\\
		\hline
	\end{tabular}
\end{table*}
\subsection{Generating driving intention input}
\label{sect:intention_images}
Driving intention describes the knowledge about what direction the vehicle will take when multiple options are available, such as, for example, at intersections or highway exits.
Intention is an important component of driving considering that people rarely drive in a purely reactive fashion and indeed adapt their driving behavior according to their destination.

Having access to the driving intention is also important when training the FCN in order to make sense of otherwise ambiguous situations: 
For example, in certain cases, when approaching an intersection the vehicle will turn right, while in others it will go straight or turn left. 
Without knowing where the driver intends to go, these examples will provide conflicting feedback and they might deteriorate the FCN's ability to learn a robust representation of the driver's model. 

In this work, Google Maps is used to obtain driving instructions when approaching locations where multiple direction can be taken. 
When queried given the current position and a certain destination, Google Maps returns a human interpretable driving action, such as \textit{turn left} or \textit{take exit}, together with an approximate distance of where that action should be taken. 
This information is integrated into the past path $\Pi^{-}$ described in Section~\ref{sect:gpsimu_images}, that is therefore augmented with two additional dimensions: \textit{intention direction}, $i_{d} \in \{\mbox{left},\mbox{straight},\mbox{right}\}$, and \textit{intention proximity}, $i_{p} \in [0, 1]$. See Fig.~\ref{fig:overviewIO} (D) and (H) for an example of intention proximity and intention direction tensors, respectively.
The default direction is to go straight, which should not be literally interpreted as keeping a constant heading angle, but simply as to keep driving along the main road, which obviously may not be straight.
The GPS positions and intention proximity are not exact and their purpose is not to tell the FCN \textit{precisely} where and how it is supposed to turn. 
They just provide an indication of the coming action that should be acted upon only if there is agreement with the scene understanding provided by the LIDAR. 

\subsection{FCN architecture}
\label{sect:architectures}
In this work, path generation is cast as a binary pixel-level semantic segmentation problem within a deep learning framework: An FCN is trained to assign to each RoI's grid cell, or pixel, a probability that the vehicle will drive over that area during its future motion. The future path is then obtained by considering the region defined by the grid cells with probability greater than a fixed threshold.

In recent years, several CNNs for carrying out semantic segmentation have been proposed; some examples are Segnet \cite{BadrinarayananEtAl2015}, FCN-8s \cite{LongEtAl2015}, and Dilation \cite{YuEtAl2015}. 
Here, however, it was preferred to implement a task-specific FCN by taking into account two factors: (1) the nature of our training data, and (2) recent design guidelines regarding semantic segmentation networks. 
In \cite{WuEtAl2016}, the authors have shown that working with high-resolution feature maps is helpful for achieving better performance.
However, the higher the resolution, the larger the memory requirements of the network. Here, to compromise between the two,  only two max-pooling layers were used. 
Additionally, the proposed FCN was designed to have a large receptive field which has also been shown to be beneficial for semantic segmentation.
The expansion of the receptive field was efficiently accomplished by using dilated convolutions \cite{YuEtAl2015}. The FCN's overall architecture is shown in Fig.~\ref{fig:network} and  further details are provided in Table~\ref{table:nets_properties}.

\section{Data set and training}
\label{sect:dataAndTraining}
The data set used in this work is entirely based on the KITTI raw data set \cite{GeigerEtAl2013}, which consists of 55 driving sequences taken over 4 days in three driving environments: city, rural, and highway. 
The sequences have lengths ranging from a few seconds to a few minutes. 
Out of the 55 available, only 45 were used: 30 were assigned to the training set, and 15 to the validation and test sets. 
Three behaviors determined by the vehicle yaw rate are defined in order to gain a coarse insight of the driving actions carried out in the sequences:  \textit{turning left} if the yaw rate is greater than $1.0^{\circ}$/s, \textit{turning right} for a yaw rate less than  $-1.0^{\circ}$/s, and \textit{straight} in all other cases \cite{XuEtAl2016}. 
A break-down of the sets is provided in Table~\ref{table:kitti_dataset}.

The FCNs were trained using the Adam optimization algorithm with an initial learning rate of 0.0005, a batch-size of 2, and using cross-entropy loss as the objective function. 
The learning rate was decayed by a factor of 2 whenever there was no improvement of performance within the last epoch. For regularization, spatial dropout layers ($p_{d}=0.20$) were added after each dilated convolution layer in the context module. Furthermore, data augmentation was carried out on-the-fly in the form of random rotations in the range $[-20^{\circ}, 20^{\circ}]$ about the LIDAR $z$\nobreakdash-axis. 
The FCNs were implemented using Torch7 framework and were trained on an NVIDIA GTX980Ti GPU. 

\begin{table}[h!]
	\centering
	\caption{Summary of the data set splits and number of examples assigned to each coarse driving action.} 
	\label{table:kitti_dataset}
	\begin{tabular}{|c|c|c|c|c|}
		\hline
		\multicolumn{1}{|c|}{Split}&\multicolumn{1}{c|}{Left}&\multicolumn{1}{c|}{Straight}&\multicolumn{1}{c|}{Right}&\multicolumn{1}{c|}{Total} \\
		\hline		
		Training & 2982  & 5836 & 3033 & 11851 \\
		Validation & 158 & 393 & 277 & 828 \\
		Test & 691 & 1503 & 1120 & 3314 \\
		\hline
	\end{tabular}
\end{table}

\section{Experiments}
\label{sect:experiments}
The following experiments had two main goals: To investigate whether the proposed approach could learn to generate feasible driving paths from real-world driving sequences, and to study how different sensor and information modalities affect performance.
For these purposes, the FCN described in Section~\ref{sect:architectures} was trained using five different combinations of input data\footnote{Videos of the FCNs applied to full driving sequences of the validation and test sets can be found at http://goo.gl/ksRrYA}: Lidar-IMU-INT(ention), Lidar-INT, Lidar-IMU, Lidar-only, and  IMU-only. IMU-only denotes the input tensor consisting of three channels: forward speed, forward acceleration, and yaw rate (see also Section~\ref{sect:gpsimu_images}). Additionally, a baseline denoted as \textit{Straight} was also included in the comparison:  this consists of a straight path of width 1.80 meters originating in the vehicle and heading forward. 
The metrics used for evaluation are precision (PRE), recall (REC), and maximum F1-measure (MaxF). In the following, we will refer to the FCNs using their corresponding input tensor description, so, for example, Lidar-IMU-INT will denote the FCN trained with LIDAR, IMU, and intention data. 

\begin{figure*}[h]
	\centering
	\includegraphics[width=2\columnwidth]{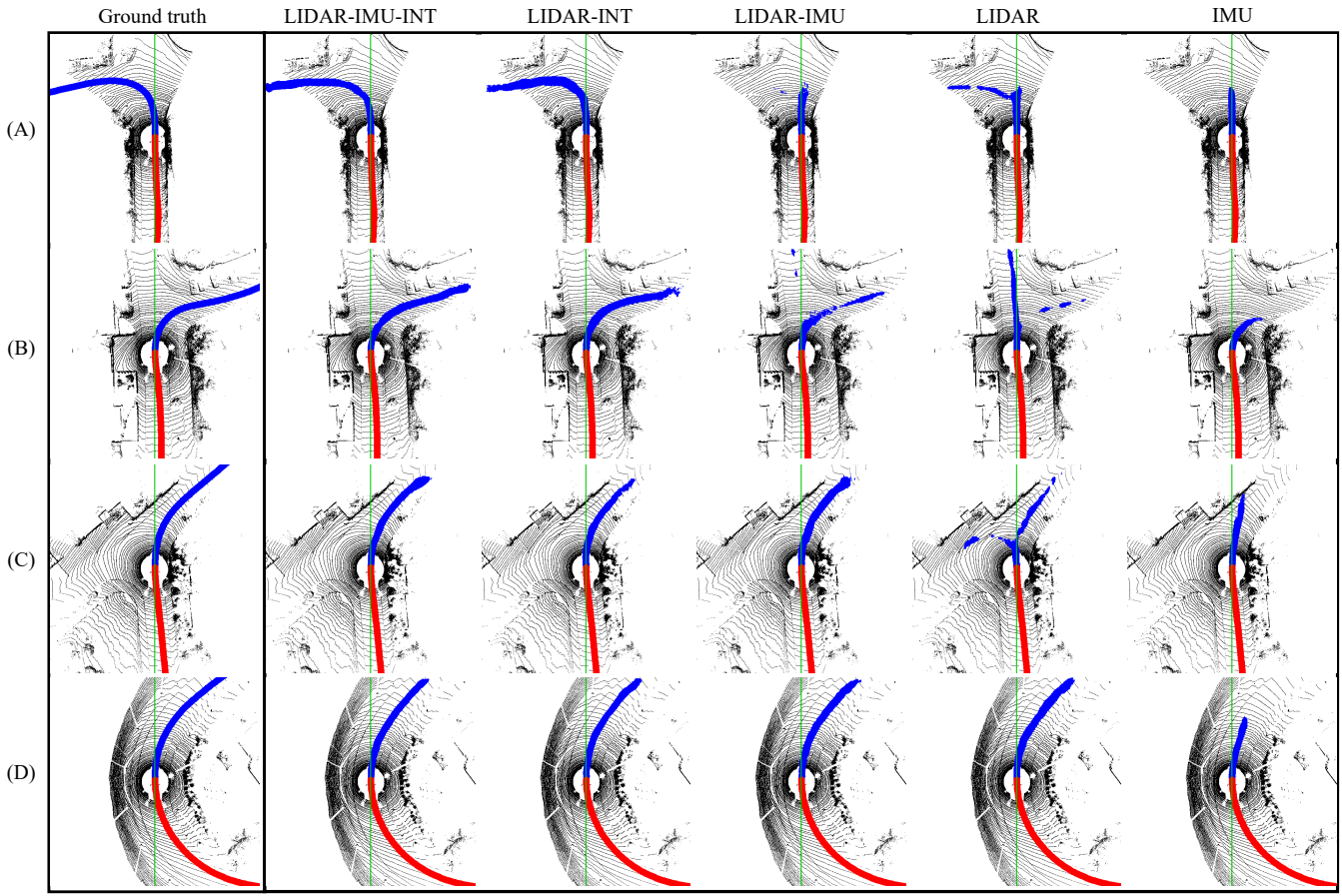}
	\caption{Qualitative comparison of the FCN's performance using several combinations of sensor and information modalities. The blue region represents the future path whereas the red region denotes the past path. (A) and (B) illustrate two cases where intention information was particularly useful: As can be seen, only Lidar-IMU-INT and Lidar-INT were able to predict accurate paths. (C) shows that IMU information can be helpful to predict the future path, especially when a turning maneuver is recently initiated. (D) shows a simple scenario consisting of a single road ahead; with the exception of IMU-only, all the other FCNs predicted accurate paths in this case.}
	\label{fig:comparison}
\end{figure*}

\begin{figure*}[h]
	\centering
	\includegraphics[width=1.85\columnwidth]{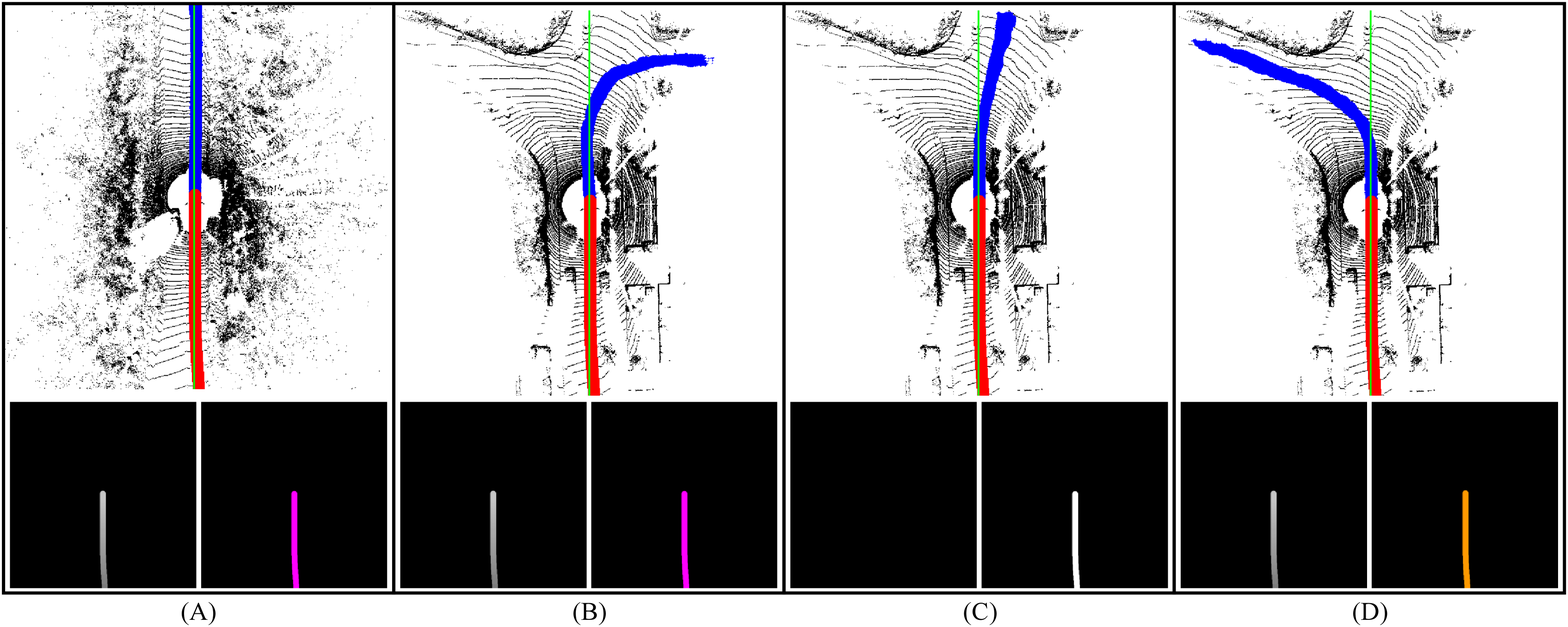}
	\caption{In each column, the top panel shows the occupancy top-view overlayed with the past path (in red) and the Lidar-IMU-INT's future path prediction (in blue). The bottom panels are the driving intention proximity (left) and driving intention direction (right). Column A illustrates an example where there is disagreement between the driving intention (turn right) and the LIDAR point cloud that shows a straight road with no exits. The system is able to determine that the driving intention cannot be satisfied and generates a correct straight path. Columns B--D show a different scenario where multiple directions are possible. By only changing the driving intention, Lidar-IMU-INT is able to generate paths that satisfy both intention and drivability.}
	\label{fig:intention}
\end{figure*}

\subsection{Results overview}
As can be seen in Table~\ref{table:kitti_results}, IMU-only obtained the lowest MaxF score of 81.63\%. 
Performance increased every time an information modality was added, reaching the highest MaxF score of 88.13\% in the case of Lidar-IMU-INT. 
These results support the intuitive assumption that the more information the FCN has access to, the better it can learn and perform the task. Furthermore, they confirm that the data representation adopted in this work (see Section~\ref{sect:overview}) is appropriate for carrying out information fusion. The average inference time, that is, the time for a forward pass through the FCN, from input to output, was 0.033 seconds, which corresponds to a frame-rate of about 30 Hz. 

The FCN was able to assign a high probability to grid cells with no LIDAR readings, such as regions falling in-between LIDAR layers or occluded areas, by exploiting context information; see row (A) of Fig.~\ref{fig:comparison}, for an example of inference within an occluded region generated by Lidar-IMU-INT.
In some cases, however, it was noticed that the predicted paths became less accurate at longer ranges. This could be because the densities of LIDAR point clouds decrease with distance from the sensor so that their information content also decreases following the same pattern. 
By considering a smaller output RoI of $40\times40$ meters, accuracy increased for all the considered input combinations; also in this case, the best performance was achieved by Lidar-IMU-INT with a MaxF score of 92.60\%.

\begin{table}[t]
	\centering
	\caption{Performance comparison on the test set considering the straight-path baseline and five input combinations. The output RoI size is specified in the first column.}
	\label{table:kitti_results}
	\begin{tabular}{|r|r|c|c|c|}
		\cline{3-5}    \multicolumn{2}{c|}{\multirow{2}[2]{*}{}} & MaxF  & PRE   & REC \\[2pt]
		\multicolumn{2}{c|}{} & \%    & \%    & \%  \\[2pt]
		\hline
		\multirow{6}[8]{*}{\rotatebox[origin=r]{90}{ \hspace{0.5cm} $60\times60$ m }} & Straight   &  68.56     & 65.57      & 71.83\\[1pt]
		& IMU-only   &  81.63     &    77.33   &  86.42   \\[2pt]
		& Lidar-only &   83.44    &   85.68    &   81.31 \\[2pt]
		& Lidar-IMU &  85.16     &   85.60    &  84.72  \\[2pt]
		& Lidar-INT &   85.76    &  88.77     &  82.94  \\[2pt]
		& Lidar-IMU-INT &  88.13     &   88.56    & 87.81  \\[2pt]
		\hline
		\multirow{6}[8]{*}{\rotatebox[origin=r]{90}{ \hspace{0.5cm} $40\times40$ m  }}& Straight   &  77.33    &  73.67     &  81.37 \\[2pt]
		& IMU-only   &   89.74    & 88.45      &   91.07  \\[2pt]
		& Lidar-only &   89.05    &   89.54    &   88.56   \\[2pt]
		& Lidar-IMU & 90.30      &   91.45    & 89.18      \\[2pt]
		& Lidar-INT &   90.76    &  91.49     &  90.04     \\[2pt]
		& Lidar-IMU-INT &   92.60    &  93.44    &   91.77   \\[2pt]
		\hline
	\end{tabular}%
\end{table}%

\subsection{Driving intention}
In Section~\ref{sect:intention_images}, it was argued that considering driving intention information would enable the FCN to learn a more robust driving model. This was indeed confirmed by the above results. Rows (A) and (B) in Fig.~\ref{fig:comparison}  illustrate two driving scenarios, in proximity of an intersection, where the benefit provided by intention information is particularly evident. As can be seen, Lidar-IMU-INT and LIDAR-INT were able to generate paths that are close to the ground-truth, whereas, the other FCNs produced uncertain predictions. It is worthwhile to mention that the driving intention only provides information about the coarse action to be executed and an approximate distance to the point where the maneuver should take place. In addition to that, the FCN must carry out scene understanding from the LIDAR point cloud in order to generate a path that, besides following the driving intention, also takes into account drivability. The intention should be acted upon only if there is agreement with the scene understanding, otherwise the system should trust the latter. Column (A) in Fig.~\ref{fig:intention} shows an example where the driver intention was to turn right even though the road did not allow such a maneuver. The system recognized that turning was not feasible and generated a correct straight path. Panels B--D illustrate the paths generated by Lidar-IMU-INT in a situation where multiple directions are possible: only the driving intention was modified whereas the LIDAR and IMU tensors were left unchanged. 

\section{Conclusion and future work}
\label{sect:conclusion}
In this work, an FCN is trained end-to-end to generate driving paths by integrating multiple sensor and information sources, that is, LIDAR point clouds, GPS coordinates, driving directions, and inertial measurements. The system generates interpretable outputs and preserves the decoupling between control and perception. Given that its training data is obtained automatically, a large volume of training examples for supervised learning could be collected with minimal effort.

Several issues are left for future work: Of particular interest to the authors is to explore approaches for integrating LIDAR point clouds acquired over successive time steps in order to generate paths that take into account the motion of nearby vehicles. Considering additional sensors such as, for example, radars and cameras could further enhance the system accuracy and perception range. The accuracy of the ground-truth paths could be improved by performing dead-reckoning in addition to using GPS coordinates. Lastly, the output of the proposed system is a probability map of future vehicle positions and it still remains to be determined how to make best use of this information for carrying out trajectory planning or vehicle control.

\section*{Acknowledgment}
The authors gratefully acknowledge financial support from Vinnova/FFI.

\bibliographystyle{IEEEtran}
\bibliography{IEEEabrv,bibliography}

\begin{thebibliography}{10}
\providecommand{\url}[1]{#1}
\csname url@samestyle\endcsname
\providecommand{\newblock}{\relax}
\providecommand{\bibinfo}[2]{#2}
\providecommand{\BIBentrySTDinterwordspacing}{\spaceskip=0pt\relax}
\providecommand{\BIBentryALTinterwordstretchfactor}{4}
\providecommand{\BIBentryALTinterwordspacing}{\spaceskip=\fontdimen2\font plus
\BIBentryALTinterwordstretchfactor\fontdimen3\font minus
  \fontdimen4\font\relax}
\providecommand{\BIBforeignlanguage}[2]{{%
\expandafter\ifx\csname l@#1\endcsname\relax
\typeout{** WARNING: IEEEtran.bst: No hyphenation pattern has been}%
\typeout{** loaded for the language `#1'. Using the pattern for}%
\typeout{** the default language instead.}%
\else
\language=\csname l@#1\endcsname
\fi
#2}}
\providecommand{\BIBdecl}{\relax}
\BIBdecl

\bibitem{ChenEtAl2015}
C.~Chen, A.~Seff, A.~Kornhauser, and J.~Xiao, ``Deepdriving: Learning
  affordance for direct perception in autonomous driving,'' in
  \emph{Proceedings of the IEEE International Conference on Computer Vision},
  2015.

\bibitem{Pomerleau1989}
D.~A. Pomerleau, ``Alvinn: An autonomous land vehicle in a neural network,'' in
  \emph{Advances in Neural Information Processing Systems 1}, D.~S. Touretzky,
  Ed.\hskip 1em plus 0.5em minus 0.4em\relax Morgan-Kaufmann, 1989, pp.
  305--313.

\bibitem{BojarskiEtAl2016}
M.~Bojarski, D.~Del~Testa, D.~Dworakowski \emph{et~al.}, ``End to end learning
  for self-driving cars,'' \emph{arXiv preprint arXiv:1604.07316}, 2016.

\bibitem{LecunEtAl2015}
Y.~LeCun, Y.~Bengio, and G.~Hinton, ``Deep learning,'' \emph{Nature}, vol. 521,
  no. 7553, pp. 436--444, 2015.

\bibitem{XuEtAl2016}
H.~Xu, Y.~Gao, F.~Yu, and T.~Darrell, ``End-to-end learning of driving models
  from large-scale video datasets,'' \emph{arXiv preprint arXiv:1612.01079},
  2016.

\bibitem{CaltagironeEtAl2017}
L.~Caltagirone, S.~Scheidegger, L.~Svensson, and M.~Wahde, ``Fast lidar-based
  road detection using convolutional neural networks,'' \emph{arXiv preprint
  arXiv:1703.03613}, 2017.

\bibitem{BadrinarayananEtAl2015}
V.~Badrinarayanan, A.~Handa, and R.~Cipolla, ``Segnet: A deep convolutional
  encoder-decoder architecture for robust semantic pixel-wise labelling,''
  \emph{arXiv preprint arXiv:1505.07293}, 2015.

\bibitem{LongEtAl2015}
J.~Long, E.~Shelhamer, and T.~Darrell, ``Fully convolutional networks for
  semantic segmentation,'' in \emph{Proceedings of the IEEE Conference on
  Computer Vision and Pattern Recognition}, 2015, pp. 3431--3440.

\bibitem{YuEtAl2015}
F.~Yu and V.~Koltun, ``Multi-scale context aggregation by dilated
  convolutions,'' in \emph{ICLR}, 2016.

\bibitem{WuEtAl2016}
Z.~Wu, C.~Shen, and A.~van~den Hengel, ``High-performance semantic segmentation
  using very deep fully convolutional networks,'' \emph{CoRR}, vol.
  abs/1604.04339, 2016.

\bibitem{GeigerEtAl2013}
A.~Geiger, P.~Lenz, C.~Stiller, and R.~Urtasun, ``Vision meets robotics: The
  kitti dataset,'' \emph{International Journal of Robotics Research (IJRR)},
  2013.

\end{thebibliography}
\end{document}